\documentclass[letterpaper, 10 pt, journal, twoside, final]{ieeeconf}  %

\IEEEoverridecommandlockouts                              %
\usepackage{multicol}
\usepackage[bookmarks=true]{hyperref}
\usepackage{comment}
\usepackage{amssymb}
\usepackage{amsmath}
\usepackage{multirow}
\usepackage{flushend}
\usepackage[dvipsnames]{xcolor}
\usepackage{pifont}
\usepackage[detect-weight=true, detect-family=true,detect-inline-weight=math]{siunitx}
\usepackage[natbib=true,
    style=ieee,
    citestyle=numeric-comp,
    backend=biber,
    maxbibnames=10,
    maxcitenames=99,
    doi=false,
    url=false,
    giveninits=true]{biblatex}
\renewcommand*{\bibfont}{\footnotesize}

\usepackage[capitalise]{cleveref} %
\crefname{equation}{}{} %

\usepackage{tikz}
\usetikzlibrary{fit,calc}
\newcommand{\Fd}{\mathbf{F}_d}
\newcommand{\Md}{\mathbf{M}_d}
\newcommand{\Ro}{\mathbf{R}_0}
\newcommand{\Rot}{\mathbf{R}_0^T}
\newcommand{\Rodot}{\Dot{\mathbf{R}}_0}

\newcommand{\mud}{\boldsymbol \mu_{i_d}}
\newcommand{\mudj}{\boldsymbol \mu_{j_d}}
\newcommand{\mubold}{\boldsymbol \mu}
\newcommand{\normmu}{\|\boldsymbol \mu_{i_d}\|}

\newcommand{\qid}{\mathbf{q}_{i_d}}
\newcommand{\qiddot}{\Dot{\mathbf{q}}_{i_d}}
\newcommand{\qidot}{\Dot{\mathbf{q}}_i}
\newcommand{\qi}{\mathbf{q}_i}

\newcommand{\muId}{\boldsymbol \mu_{1_d}} %
\newcommand{\mund}{\boldsymbol \mu_{n_d}}
\newcommand{\nij}{\mathbf{n}_{ij}}
\newcommand{\nii}{\mathbf{n}_{i}}
\newcommand{\nj}{\mathbf{n}_{j}}
\newcommand{\normnij}{\|\mathbf{n}_{ij}\|}

\newcommand{\opi}{\mathbf{p}_i}
\newcommand{\opj}{\mathbf{p}_j}
\newcommand{\po}{\mathbf{p}_0}

\newcommand{\hatpai}{\hat{\mathbf{p}}_{a_i}}
\newcommand{\hatpaj}{\hat{\mathbf{p}}_{a_j}}
\newcommand{\pai}{\mathbf{p}_{a_i}}
\newcommand{\paj}{\mathbf{p}_{a_j}}

\newcommand{\Fdi}{\mathbf{F}_{d_i}}
\newcommand{\Fdj}{\mathbf{F}_{d_j}}

\newcommand{\quati}{\mathbf{q}_i}
\newcommand{\quatj}{\mathbf{q}_j}

\newcommand{\QPmu}{\text{QP}_{\boldsymbol\mu}}
\newcommand{\QPsvm}{\text{QP}_{svm}}
\newcommand{\QPFd}{\text{QP}_{\Fd}}

\title{\LARGE \bf
Efficient Optimization-based Cable Force Allocation for Geometric Control of a Multirotor Team Transporting a Payload
}

\author{Khaled Wahba and Wolfgang Hönig
\thanks{Manuscript received: August 11, 2023; Revised: November 11, 2023; Accepted: December 11, 2023. This paper was recommended for publication by Editor M. Ani Hsieh upon evaluation of the Associate Editor and Reviewers’ comments.}
\thanks{All authors are with Faculty of Electrical Engineering and Computer Science,
        Technical University of Berlin, Berlin, Germany
        {\tt\footnotesize \{k.wahba, hoenig\}@tu-berlin.de}.}%
\thanks{The work was supported by the Deutsche Forschungsgemeinschaft (DFG,
German Research Foundation) - 448549715.}
\thanks{Video: \url{https://youtu.be/3OfnUIiu3bs}}%
\thanks{Code: \url{https://github.com/IMRCLab/col-trans}}%
\thanks{Digital Object Identifier (DOI): see top of this page.}
}

\begin{document}

\maketitle
\begin{abstract}

We consider transporting a heavy payload that is attached to multiple multirotors.
The current state-of-the-art controllers either do not avoid inter-robot collision at all, leading to crashes when tasked with carrying payloads that are small in size compared to the cable lengths, or use computational demanding nonlinear optimization.
We propose an efficient optimization-based cable force allocation for a geometric payload transport controller to effectively avoid such collisions, while retaining the stability properties of the geometric controller.
Our approach introduces a cascade of carefully designed quadratic programs that can be solved efficiently on highly constrained embedded flight controllers.

We show that our approach exceeds the state-of-the-art controllers in terms of scalability by at least an order of magnitude for up to 10 robots. We demonstrate our method on challenging scenarios with up to three small multirotors with various payloads and cable lengths, where our controller runs in realtime directly on a microcontroller on the robots.

\end{abstract}
\section{Introduction}

Aerial vehicles can access remote areas, making them suitable for collaborative tasks at a construction site, rubble removal in search-and-rescue scenarios, or nuclear power plant decommissioning. 
Cable-driven payload transportation using multi-UAVs is beneficial as it avoids carrying manipulators or grippers onboard, enabling the transportation of heavier objects~\cite{masone2016cooperative}, especially tools or supplies~\cite{gabellieri2018study}.

State-of-the-art controllers for a team of multirotors to transport a cable-suspended payload include geometric controllers~\cite{lee2017geometric,lee2013geometric} as well as nonlinear model predictive control~\cite{li2023nonlinear, sun2023nonlinear}.
These existing works either ignore possible collisions between robots~\cite{lee2017geometric}, or use a nonlinear optimization framework that necessitates complex on-board computation~\cite{li2023nonlinear, liu2022safety}.
Collision between robots can easily occur for payloads that are small in size compared to the robot's size and the cable length which poses a safety risk. 

\begin{figure}
    \centering
    \includegraphics[width=0.49\linewidth]{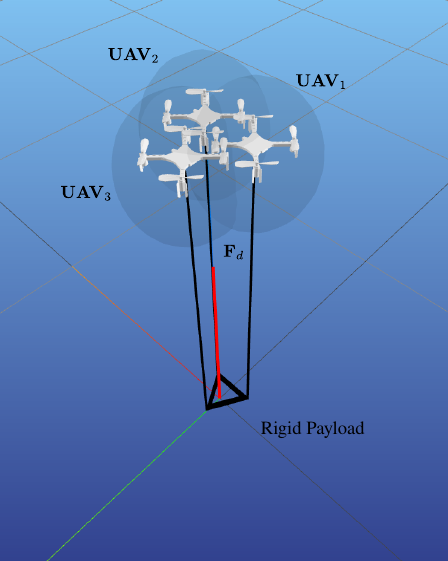}
    \includegraphics[width=0.49\linewidth]{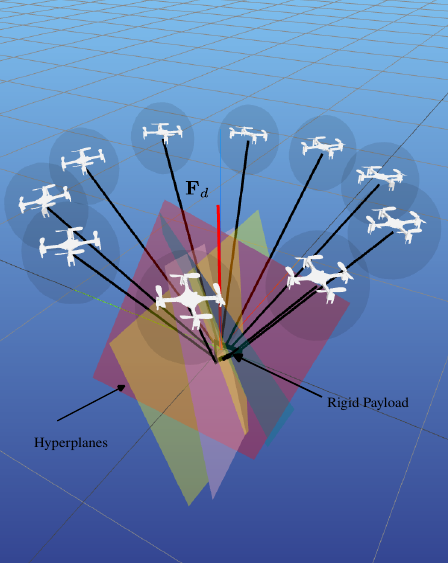}

    \caption{
    We contribute an efficient payload-type-agnostic approach to compute desired cable forces that avoids collisions.
    Left: Three multirotors carrying a rigid body payload with the baseline controller~\cite{lee2017geometric} . The desired cable forces are allocated with \cref{eq:baseline}, causing inter-robot collisions.
    Right: Our approach with computed hyperplanes that define a safe convex set for the cable forces, while considering the desired motion of the payload.
    }
    \label{fig:overview}
\end{figure}

We propose a new distributed cable force allocation method for the geometric controller by \cite{lee2017geometric} that provides the team with new capabilities, see \cref{fig:overview}. 
In particular, we directly consider safety distances between robots and allow for user-preferred formations.
Our approach unifies the handling of point mass~\cite{lee2013geometric} and rigid body~\cite{lee2017geometric} payload types.
We use a cascaded design of three low-dimensional quadratic programs (QPs), which have global minima that can be found within milliseconds even on highly resource-limited microcontrollers.
In fact, this approach allows us to execute the control law in a distributed fashion on physical robots.
Effectively, our method allows users to teleoperate a team of robots by simply commanding the payload, independent of the payload size or type.
Our contributions are as follows:
\begin{enumerate}
    \item Algorithmically, we provide a novel distributed and payload-type-agnostic QP-based force allocator that can be deployed on microcontrollers.
    \item Empirically, we demonstrate our distributed controller on a team of up to three small (\SI{34}{g}) multirotors carrying different payloads. We also show a superior scalability with the number of robots compared to existing state-of-the-art controllers.
\end{enumerate}

We note that we are the first to demonstrate the geometric controller on a highly compute-constrained flight controller.

\section{Related Work}
The dynamics of the cable-suspended payload system with multi-UAVs  can be described explicitly through the interaction forces between the payload, cables, and the UAVs~\cite{masone2016cooperative}. In general, the dynamics are expressed using Newton's equations of motion~\cite{pereira2017control,tognon2018aerial,tuci2018cooperative}, while augmenting the rotational dynamics of the UAVs~\cite{lee2017geometric}.

There has been some advancement in simulation, planning, and control of payload collaborative transport. One method involves simulating the system dynamics \cite{li2022rotortm} and uses offline motion planners \cite{manubens2013motion, de2019flexible} to generate references for states, sacrificing reactivity during offline planning.

Conversely, \textbf{centralized controllers} have been devised for the entire system in a cascaded reactive fashion~\cite{sreenath2013dynamics, lee2017geometric,lee2013geometric, six2017kinematics}. These are based on force or kinematic analysis and often ensure exponential stability. However, practical challenges remain, such as the need to measure noisy payload accelerations. Estimates using disturbance observers~\cite{zhao2023composite} are tested in simulation, only. Moreover, these methods do not take inter-robot collisions and cable tangling into account.

\textbf{Decentralized controllers} enhance robustness and scalability \cite{jackson2020scalable, sharma2023decentralized}, but have only been tested in simulations or assume the presence of reliable communication.  
Other methods do not rely on communication \cite{tognon2018aerial, tagliabue2019robust}. Nevertheless, inter-robot collisions are not considered by these methods.

\textbf{Optimization-based controllers} may use iterative gradient-based solvers, but they are prone to local minima \cite{jimenez2022precise,petitti2020inertial}. Another approach is to constrain the cables into predefined convex regions \cite{geng2020cooperative}. However, all these methods lack direct consideration of inter-robot collisions.
Nonlinear model predictive control (NMPC) can control the payload and avoid inter-robot collisions \cite{li2023nonlinear, sun2023nonlinear}. However, these methods are either limited to simulations or computationally costly for highly constrained microcontrollers, and scalability with the number of robots remains poor. %

The closest methods to our approach allocate cable forces by employing either null-space methods with closed-form solutions~\cite{liu2022safety} or nonlinear optimization~\cite{liu2022safety, li2014coordinated}, including inter-robot collisions.
In contrast, our method uses convex optimization and supports novel use cases, such as different cable lengths, incorporating load distribution and inter-robot collision constraints given a desired formation. 
Thus, these two prior methods are unsuitable for direct comparison to our work.
Our approach utilizes a cascade of solving consecutive convex programs where constraints are defined by cutting off parts of the non-convex solution in exchange of finding a solution quicker. Thus, our method is not equivalent to any existing non-convex formulation \cite{liu2022safety, li2014coordinated}. 
We note that we have not seen cases where our convexification caused infeasibility.
On the other hand, non-convex formulations might fail without a good initial guess.

This paper builds on existing controllers \cite{lee2017geometric, lee2013geometric} and enhances them by incorporating consecutive quadratic programs to optimize the cable force allocation that take into account various constraints for different use cases while retaining stability. 
Moreover, we demonstrate that our method surpasses the state-of-the-art controller \cite{sun2023nonlinear} in terms of scalability and computational costs. It is noteworthy that \cite{sun2023nonlinear} was the only work that provided a detailed scalability analysis.

\section{Background}
\begin{figure*}
    \centering
    \includegraphics[width=\linewidth]{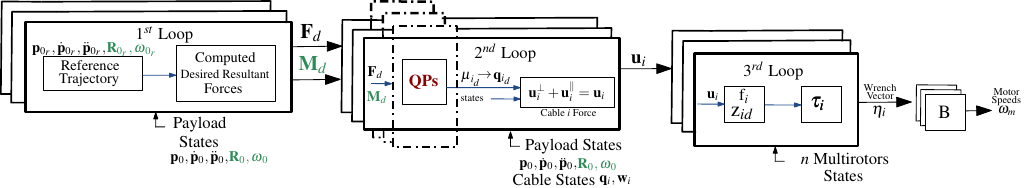}
    \caption{Controller architecture of the state-of-the-art geometric controller for payload transport. We contribute an optimization-based cable force allocation (shown in red). Green states are only needed when transporting a rigid body rather than a point mass.}
    \label{fig:fullarch}
\end{figure*}

\subsection{Single Multirotor Dynamics}
The dynamics of a single multirotor is modeled as a 6 degrees-of-freedom floating rigid body with mass $m$ and diagonal moment of inertia $\mathbf{J}$. The multirotor's state comprises of the global position $\mathbf{p} \in \mathbb{R}^3$, global velocity $\mathbf{v} \in \mathbb{R}^3$, attitude rotation matrix $\mathbf{R} \in SO(3)$ and body angular velocity $\boldsymbol{\omega} \in \mathbb{R}^3$. The dynamics can be expressed using Newton-Euler~\cite{mellinger2011minimum} equations of motion as follows
\begin{subequations}
    \begin{align}
        &\Dot{\mathbf{p}} = \mathbf{v}, && m\mathbf{\dot{v}} = m\mathbf{g} + \mathbf{R}\mathbf{f}_u, \label{eq:translationquad}\\
        &\Dot{\mathbf{R}} = \mathbf{R}\boldsymbol{\hat{\omega}}, &&
        \mathbf{J}\Dot{\boldsymbol{\omega}} = \mathbf{J}\boldsymbol{\omega}\times \boldsymbol{\omega} + \boldsymbol{\tau}_u,
    \end{align}
\end{subequations}
where $\hat{\boldsymbol{\cdot}}$ denotes a skew-symmetric mapping $\mathbb R^3 \rightarrow \mathfrak{s}\mathfrak{o}(3)$; $\mathbf{g} = (0,0,-\emph{g})^T$ is the gravity vector; $
\boldsymbol{f}_u = (0,0,f)^T$ and $\boldsymbol{\tau}_u = (\tau_x, \tau_y, \tau_z)^T$ are the total thrust and body torques from the rotors, respectively. 
The total wrench vector applied on the center-of-mass (CoM) of the multirotor's body is defined as $\boldsymbol{\eta} = (f,\tau_x,\tau_y,\tau_z)^T$. The total wrench vector $\boldsymbol{\eta}$ is linearly related to the squared motor rotational rate (i.e., propeller speed) for $k$ motors $\boldsymbol{\omega_m} = (w_1^2,\hdots,w_k^2)^T$, such that $\boldsymbol{\eta} = \mathbf{B}_0\boldsymbol{\omega_m}$, where $\mathbf{B}_0$ is the actuation matrix~\cite{mellinger2011minimum}.

\subsection{Full System Dynamics} \label{subsec:fulldynamics}
Consider a team of \emph{n} multirotors that are connected to a payload through massless, and non-extensible cables (see \cref{fig:overview}). In the following, we summarize results presented in~\cite{lee2017geometric, lee2013geometric} using our own notation for clarity. Throughout this work, the variables related to the payload are denoted by the subscript $0$, and the variables for the \emph{i}-th multirotor are denoted by the subscript $i \in \{1,\hdots,n\}$.
The payload is described as a rigid body with mass $m_0$ and moment of inertia matrix $\mathbf{J}_0$.
The cables are assumed to be always taut (i.e., modeled as rigid rods) each with length $l_i$.
The states can be described as the global position $\mathbf{p}_0 \in \mathbb{R}^3$ and the velocity $\Dot{\mathbf{p}}_0 \in \mathbb{R}^3$ of the payload; the attitude rotation matrix $\Ro \in SO(3)$ and the body angular velocity $\boldsymbol{w}_0 \in \mathbb{R}^3$ of the payload; the unit directional vector $\qi \in \mathbb{R}^3$ and the angular velocity $\boldsymbol{\omega}_i \in \mathbb{R}^3$ of each \emph{i}-th cable, where $\qi$ points from multirotor \emph{i} towards the payload.
Hence, the state space has dimension $13+6n$.

Let us define the attachment point vectors between each cable and the payload as ${}^0\pai$ expressed in the payload reference frame. Given the states of the full system, the position of each multirotor expressed in the fixed global frame is 
\begin{equation} \label{eq:uavpos}
    \mathbf{p}_i = \mathbf{p}_0 + \Ro{}^0\pai - l_i\qi. 
 \end{equation}
Thus, the kinematics of the full system is described as 
\begin{align}
    &\qidot = \boldsymbol{\omega}_i \times \qi, & 
    \Dot{\mathbf{p}}_i = \Dot{\mathbf{p}}_0 +\Rodot{}^0\pai - l_i\qidot \nonumber\\
     &\Rodot = \Ro\boldsymbol{\hat{\omega}}_0,
\end{align}
and the nonlinear dynamics of the full system can be expressed using Euler-Lagrange equations \cite{lee2013geometric, lee2017geometric}. 

\subsection{Control Design}
\label{sec:ctrl}
\subsubsection{Single Multirotor without Payload} \label{sec:ctrlUAV}
The goal for a single multirotor controller is to compute propeller speeds such that a given a tuple reference trajectory $\langle \mathbf{p}_{r}, \Dot{\mathbf{p}}_{r}, \Ddot{\mathbf{p}}_{r}, \boldsymbol{\psi}_{r} \rangle$ for the multirotor CoM is tracked.
Here, $\mathbf{p}_r$ and $\boldsymbol{\psi}_r$ are the position and the heading of the multirotor, respectively.
The controller in~\cite{lee2010geometric} consists of a 2-\emph{loop} cascaded design, where the outer loop computes the desired force control vector, which then computes $\emph{f}$ and the desired third body \emph{z}-axis $z_d$. Thus, using differential flatness along with the reference heading $\boldsymbol{\psi}_{r}$, the desired rotational states are computed and $\boldsymbol{\tau}_{u}$ is used to track those desired states~\cite{mellinger2011minimum, lee2010geometric}. 

\subsubsection{Control Design for Rigid Payload Dynamics} \label{sec:baselinecontrol}

For the multi-UAVs payload transport system, a new controller design is needed since the single control design does not include the dynamics of the cables.

The geometric controller by \cite{lee2017geometric} solves the multi-UAVs payload transport problem, where the control problem is defined as follows. The input is a tuple reference trajectory $\langle \textbf{p}_{0_r}, \Dot{\textbf{p}}_{0_r}, \Ddot{\textbf{p}}_{0_r}, \mathbf{R}_{0_r}, \boldsymbol{\omega}_{0_r} \rangle$, where $\mathbf{p}_{0_r}, \mathbf{R}_{0_r}, \boldsymbol{\omega}_{0_r}$ are the position, rotation matrix and angular velocity of the payload, respectively. The output is a decentralized controller that computes the motor signals for each robot. 
\paragraph{Desired Payload Forces and Moments}
This controller consists of a 3-\emph{loop} cascaded design as shown in \cref{fig:fullarch}.
The first step in the first loop computes the desired resultant payload control forces, $\Fd$, and moments, $\Md$, to track the payload reference trajectory.
Then, the first loop outputs the desired cable unit directional vectors $\qid$ in order to achieve $\Fd$ and $\Md$. 
Let $\mud$ be the desired \emph{i}-th cable force and $\mathbf{P} \in \mathbb{R}^{6 \times 3n}$ be a matrix that maps the forces and torques of the payload CoM to desired cable forces $\mud$ on the attachment points $\pai$. 
Thus, $\mathbf{P}$ is a constant matrix structured from $\pai$, refer to \cite{lee2017geometric} for details.
The cable force allocation relation between $\mud$ and $\Fd, \Md$ is defined as 
\begin{equation}
    \label{eq:linearmap}
    \mathbf{P} 
    \begin{pmatrix}
    (\Rot \muId)^T 
    \hdots 
    (\Rot \mund)^T
    \end{pmatrix} = 
    \begin{pmatrix}
        (\Rot \Fd)^T &
        \Md^T
    \end{pmatrix}.
\end{equation}
\paragraph{Cable Force Allocation} \label{sec:forceAllocation}
For any $\Fd, \hspace{0.1cm} \Md$ and $n\geq 3$, there exist desired cable forces $\mud$. These desired cable forces can be computed using the Moore-Penrose inverse of $\mathbf{P}$ to achieve a least-square solution:
\begin{equation}
\label{eq:baseline}
\mud = \text{diag}(\Ro,\hdots, \Ro) \mathbf{P}^T (\mathbf{P}\mathbf{P}^T)^{-1}
\begin{pmatrix}
    \Rot \Fd \\
    \Md
\end{pmatrix}.
\end{equation}
\subsubsection{Control Design for Point Mass Dynamics}
Let us consider the special case of the payload being a point mass with $n \geq 2$, then the controller architecture retains the same structure, see \cref{fig:fullarch}. However, \cite{sreenath2013dynamics, lee2013geometric} solve the allocation of the cable forces $\mud$ by providing a user-specified team formation which is then projected on the null space of the cable forces to satisfy the following
\begin{equation}
    \label{eq:Reducedlinearmap}
    \mathbf{P} 
    \begin{pmatrix}
 \muId^T & \hdots & \mund^T
    \end{pmatrix}^T = 
     \Fd.
\end{equation}
\subsubsection{Control Inputs of the UAVs}\label{sec:controlinp}
Let the control force applied by each \emph{i}-th multirotor on its cable be $\mathbf{u}_i = \mathbf{R}_i\mathbf{f}_{u_i}$. Denote that $\mathbf{u}_i^{\parallel} \in \mathbb{R}^3$ and $\mathbf{u}_i^{\perp} \in \mathbb{R}^3$ are the orthogonal projection of $\mathbf{u}_i$ along $\qi$ and to the plane normal to $\qi$, respectively, i.e., $\mathbf{u}_i = \mathbf{u}_i^{\parallel} + \mathbf{u}_i^{\perp}$.
We compute $\qid$ using the desired cable forces $\mud$ by applying the definition 
\begin{equation}
\label{eq:desiredDirections}
    \qid = \frac{\mud}{\normmu}.
\end{equation}
The second loop computes the control force $\mathbf{u}_i$ applied by each multirotor on the payload. In particular, $\mathbf{u}_i^{\parallel}$ is first computed by projecting $\mud$ on the current cable force vector $\boldsymbol{\mu}_i$, in addition to non-linear terms that linearize the translational dynamics of the payload. In other words, when $\boldsymbol{\mu}_i \rightarrow \mud$, then $\Fd, \Md$ are tracked.
In order to track the desired cable forces $\mud$, the vector $\mathbf{u}_i^{\perp}$ is used to track the desired unit directional vector, i.e., $\qi \rightarrow \qid$, of the cables. 
When $\qi =\qid$ then the resultant force and moment acting on the payload become identical to their desired values.

After computing $\mathbf{u}_{i}$, the final third loop computes the thrust magnitude $\emph{f}$ (i.e., $\mathbf{f}_u$) and the desired third body vector $\mathbf{z}_d$. Thus the attitude control input $\boldsymbol{\tau}_u$ can be computed using $\mathbf{f}_u$ as presented in \cref{sec:ctrlUAV}.

\subsection{Quadratic Programs and Hyperplanes}
\label{sec:hp}
A quadratic program (QP) is an optimization problem with a quadratic objective and affine equality and inequality constraints. The QP is formulated as
\begin{align}\label{eq:QP}
\min_{\boldsymbol{x}} \quad &  \frac{1}{2}\boldsymbol{x}^T\boldsymbol{P}\boldsymbol{x} + \boldsymbol{q}^T\boldsymbol{x} \\
 &\text{\noindent s.t.}\begin{cases} 
\underline{\boldsymbol{u}} \leq \boldsymbol{A}\boldsymbol{x} \leq  \overline
{\boldsymbol{u}}
\nonumber 
\end{cases},
\end{align}
where $\boldsymbol{x}\in \mathbb{R}^n$ is the decision variable vector. The objective function is defined by a positive semi-definite matrix 
$\boldsymbol{P} \in \mathbb{R}^{n \times n}$ and 
$\boldsymbol{q} \in \mathbb{R}^{n}$. 
The linear constraints are defined by matrix 
$\boldsymbol{A} \in \mathbb{R}^{m \times n}$ and the lower and upper bound vectors are 
$\underline{\boldsymbol{u}}$ and 
$\overline{\boldsymbol{u}}$ respectively, such that 
$\underline{u}_i \in \mathbb{R} \cup \{-\infty\}$ and  $\overline{u}_i \in \mathbb{R} \cup \{+\infty\}$ for all $i \in \{1,\hdots,m\}$. QPs are convex and thus have a global minima.  

A hyperplane $\mathcal{H}$ in $\mathbb{R}^d$ can be formulated by a normal vector $\mathbf{n}$ and an offset \emph{a} as $\mathcal{H} = \{\boldsymbol{x} \in \mathbb{R}^d \hspace{0.1cm} | \hspace{0.1cm} \mathbf{n}^T\boldsymbol{x} - a = 0\}$. A half-space $\Tilde{\mathcal{H}}$ in $\mathbb{R}^d$ is a subset of $\mathbb{R}^d$ that is bounded by a hyperplane such that $\Tilde{\mathcal{H}} = \{\boldsymbol{x} \in \mathbb{R}^d \hspace{0.1cm} | \hspace{0.1cm} \mathbf{n}^T\boldsymbol{x} - a \leq 0\}$. The intersection of multiple hyperspaces creates a polyhedra. Hyperplanes can be used as a linear constraints for a QP formulation.

\section{Approach}
\subsection{Overview}
\begin{figure}[t]
    \centering
    \includegraphics[width=\linewidth]{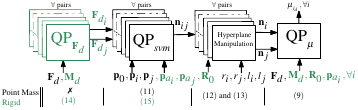}
        \caption{A cascade of three QPs allocate the desired cable forces $\mud$ for $n\geq2$ robots. The final QP, $\QPmu$, computes $\mud$ by constraining the cable forces to fulfill the allocation constraints \cref{eq:Reducedlinearmap} (point mass) or \cref{eq:linearmap} (rigid body) and limiting them to be inside a polyhedra. This volume is defined by a set of hyperspaces, which are constructed by two other QPs that are solved for each robot pair, $\QPsvm$ and $\QPFd$. 
    Black represents both payload types and green is exclusively used for the rigid payload.
    }
    \label{fig:qp_arch}
\end{figure}
We augment the existing geometric controller described in \cref{sec:ctrl} with a custom force allocation method (replacing the standard approach presented in Section \ref{sec:forceAllocation}) to achieve trajectories for the multirotors that avoid inter-robot collisions. Our approach employs a unifying framework that is capable of handling point mass and rigid payloads.
As such, we provide a drop-in replacement for \cref{eq:baseline}, that is we use the desired force on the payload, $\Fd$, and the desired moments for the payload, $\Md$, as input and compute the desired cable forces, $\mud$, while taking inter-robot collision avoidance into account. 
The nonlinear optimization problem formulation can be defined as,
\begin{align}
    &\min_{\mud} \hspace{0.2cm} \frac{1}{2}\normmu^2 \label{eq:general-optimization-problem}\\
    &\text{\noindent s.t.}\begin{cases}
        \cref{eq:linearmap} \text{ or }\cref{eq:Reducedlinearmap} \nonumber\\
       \|\opi(\mud, \boldsymbol\cdot)- \opj(\mudj, \boldsymbol\cdot)\| \geq r_i + r_j , \quad\forall i\neq j, \nonumber
    \end{cases}
\end{align}
where 
$\opi$($\mud$, $\boldsymbol\cdot$) and  $\opj$($\mudj$, $\boldsymbol\cdot$) are the positions of each robot pair (see \cref{eq:uavpos},\cref{eq:desiredDirections}) and $r_i, r_j$ are the safety radii. Consider the tracking of the desired cable directions $\qid$ is fast enough such that $\qi = \qid$. 
The first constraint enforces the force allocation constraints for the point mass or rigid body payload case, respectively.
The second constraint ensures that the safety distance between each pair of \emph{i} and \emph{j} robots is at least $r_i + r_j$.
Without the second constraint, the optimal solution for the desired cable forces is the same as in \cref{eq:baseline}.
Instead of using a nonlinear optimization that solves for $\mud$ subject to the second nonconvex constraint, we use a cascaded design of three quadratic programs (QPs) to link these variables implicitly, see \cref{fig:qp_arch}.
The first three blocks in \cref{fig:qp_arch} run in a decentralized manner on each multirotor to compute all hyperplanes for all robot pairs.
Thus, for $n$ multirotors, these blocks are executed $n(n-1)/2$ times.

Our key insight is to construct hyperplanes that separate each pair of multirotors. 
These hyperplanes should also take the desired motion of the payload (i.e., $\Fd, \Md$) into account. 
For each pair we use the first block, $\QPFd$, to provide heuristics of desired cable forces ($\Fdi$ and $\Fdj$) that approximate the actual solution to achieve $\Fd, \Md$, which will guide the hyperplane generation.
The next block, $\QPsvm$, constructs a hyperplane to separate the two corresponding cables with a preference for the approximate solutions $\Fdi$ and $\Fdj$ (i.e., rotating or offsetting the hyperplane).  
Since the resulting hyperplane between each pair has a small safety margin, we use geometric hyperplane manipulation (third block) to construct two new hyperplanes, one for each multirotor of the pair, that possess the desired safety distance and constrains the desired cable forces $\mud$.
Note that this geometric manipulation is computationally better compared to solving one optimization problem per hyperplane.

Finally, the last block $\QPmu$ runs once on each robot to allocate the desired cable forces $\mud$ for the full cable system, and each multirotor extracts its own solution from $\QPmu$.
This QP always finds the same solution for the same problem input when $n\geq2$, since it has one global minima only. 

\subsection{QP For Desired Cable Forces (\texorpdfstring{$\text{QP}_{\boldsymbol\mu}$}{QP\_mu})}
The minimum-norm solution of \cref{eq:linearmap} or \cref{eq:Reducedlinearmap} provides a solution for $\mud$. We add additional (hard) constraints to the optimization. The resulting optimization problem is a QP due to the linear constraints and can be solved efficiently:
\begin{align}
    &\min_{\mud} \hspace{0.2cm} \frac{1}{2}\|\mud\|^2 \label{eq:newqprigid}
    \\
    &\text{\noindent s.t.}\begin{cases}
    \cref{eq:linearmap} \text{ or }\cref{eq:Reducedlinearmap} \\
    \begin{pmatrix}
        \mathbf{n}_1^T & \hdots & 0 \\
        0 & \ddots & 0 \\
        0 &  \hdots  & \mathbf{n}_m^T
    \end{pmatrix}
    \begin{pmatrix}
    \muId \\
    \vdots \\
    \mund
    \end{pmatrix} - 
    \begin{pmatrix}
        a_1 \\
        \vdots \\
        a_m
    \end{pmatrix}
    \leq \mathbf{0}_{m\times1} 
    \end{cases}, \nonumber
\end{align}
where $\mathbf{n}_i \in \mathbb{R}^{3\times1}$ and $a_i \in \mathbb R$ for $i \in {1,\hdots,m}$ define the half-spaces and $m$ is the number of hyperplanes with $m \geq n$.

\subsection{Point Mass Model} \label{sec:pointmassmodel}

In the $\QPmu$ formulation, our inequality constraints are formulated using hyperplanes.
We compute these hyperplanes by solving another QP followed by geometric manipulation to construct a constraint for the cable forces.
Consider each pair of robots $i$ and $j$.
Their positions are ${}^0\opi$ and ${}^0\opj$ in the payload frame of reference.
For better readability, we omit the superscripts of the robot position vectors in the payload frame, and use them as $\opi$ and $\opj$. 

\subsubsection{{QPs for Hyperplanes} (\texorpdfstring{$\text{QP}_{svm}$}{QP\_svm})} \label{subsec:hp_gen_pm}

For each pair of robots we generate two hyperplanes $\nii$ and $\nj$, see \cref{fig:2cfs_pm} on the left.
In order to generate $\nii, \nj$, we first compute for each pair the hyperplane $\nij$ that separates both robots while being as close as possible to $\Fd$. Thus, $\nij$ maximizes the safety margin between robots positions while regularizing for $\Fd$. This regularization term guarantees that $\mud$ can be tracked well by allowing the cables to move in the same direction of $\Fd$.
Afterwards, $\nij$ is used to compute both $\nii$ and $\nj$ to be used in QP$_{\boldsymbol \mu}$. 
We adopt a variation of multi-objective hard-margin support vector machines (SVM)  with additional regularization for $\Fd$ as $\QPsvm$:
\begin{align}
    &\min_{\nij} \hspace{0.2cm} \normnij^2 + \lambda_s(\nij^T\Fd)^2
    \label{eq:svmpm}
    \\
    &\text{\noindent s.t.}\begin{cases}
    \nij^T\opi  \leq -1 , \quad
    \nij^T\opj \geq 1 \\
    \end{cases}. \nonumber
\end{align}
Here, $\lambda_s$ is a weighting factor for the soft constraint that acts as a trade-off between the safety margin and the hyperplane of $\nij$ being as close to $\Fd$. In fact, a high $\lambda_s$ might result in a safety margins that are close to zero. Our formulation implicitly constrains the plane to pass through the payload by not including an offset $a$ in the $\QPsvm$. If such an offset is included, then the hyperplanes might be shifted and consequently, this might render $\QPmu$ infeasible.

\subsubsection{Hyperplane Manipulation} \label{subsec:hyperplane_man_pm}

After computing the separating hyperplane $\nij$, we need to compute $\nii$ and $\nj$ normals of the hyperplanes as constraints for the QP in \cref{eq:newqprigid}, see \cref{fig:2cfs_pm}. It is worth noting that the $\nij$ vector is always directed towards $\opi$. 
Given the radius of each robot $r_i$ and $r_j$ and the length of each cable $l_i$ and $l_j$, we first compute the angles $ \alpha_i$ and $\alpha_j$ by using the properties of the resulting isosceles triangle as
\begin{align}
\label{eq:alphas}
    \alpha_i = 2\arcsin\left(\frac{r_i}{2l_i}\right), \quad \alpha_j = 2\arcsin\left(\frac{r_j}{2l_j}\right).
\end{align}
Let us define $\mathbf{e}_3 = (0,0,1)^T$ and define quaternions $\quati$ and $\quatj$ by converting the axis-angle representation to a quaternion using the axis $\nij \times \mathbf{e}_3$ with the angles $ \alpha_i$ and $-\alpha_j$. Finally, we compute the normals $\nii$ and $\nj$ by tilting $\nij$ with the computed $\quati$ and $\quatj$. Let us define $\odot$ as the quaternion rotation operator, then $\nii$ and $\nj$ are
\begin{align}
\label{eq:nij}
    & \nii = \quati \odot \nij, \quad \nj = -\quatj \odot \nij.
\end{align}
Note that it is required by $\QPmu$ in \cref{eq:newqprigid} that both normals must be pointing inwards (i.e., towards each other), which explains the negative sign added to \cref{eq:nij} for the $\nj$ normal vector computation.

\subsection{Rigid Body Model}
Similar to the point mass model (see \cref{sec:pointmassmodel}), our objective for the rigid payload is to find the hyperplanes $\nii$ and $\nj$ (\cref{fig:2cfs_pm} on the right) that set the desired safety distances between the robots.
However, in the rigid payload case the hyperplane $\nij$ (which computes $\nii$ and $\nj$) needs to not only take into account being close to $\Fd$, but also the desired moments $\Md$.
Thus, we propose a special QP for the rigid payload case, $\QPFd$ (see \cref{fig:qp_arch}). Here, $\QPFd$ includes $\Md$ to generate the $\nij$ hyperplane.
\subsubsection{{QPs for Force Pairs} (\texorpdfstring{$\text{QP}_{\Fd}$}{QP\_Fd})} \label{subsec:hp_gen_Fd}
We first solve for each pair of robots $\QPFd$, which computes virtual heuristic forces $\Fdi \text{ and } \Fdj$ to be applied on the attachment points $\pai$ and $\paj$. 
Recall that the attachment points of each cable are $\pai, \paj$ in the payload reference frame.
The sum of these forces must achieve $\Fd$ while producing moments close to $\Md$. 
Let us define $\hatpai$ and $\hatpaj$ as the skew-symmetric mapping of $\pai$ and $\paj$, then $\Fdi $ and $\Fdj$ can be solved as 
\begin{align} \label{eq:QPFd}
    &\min_{\Fdi, \Fdj} \|\Fdi\|^2 + \| \Fdj \|^2 + \\ &\hspace{1.1cm} \| \Md - (\hatpai\Rot\Fdi + \hatpaj\Rot\Fdj) \|^2 \nonumber 
    \\
    &\text{\noindent s.t.}\begin{cases}
        \Fdi + \Fdj = \Fd.  \nonumber
    \end{cases}
\end{align}
Since the output of $\QPFd$ are heuristic virtual forces $\Fdi$ and $\Fdj$, we do not need to track $\Md$ as a hard constraint. In particular, the mapping between $(\Fd, \Md)$ and $(\Fdi,\Fdj)$ might be infeasible depending on the attachment points positions $(\pai, \paj)$.
Thus, we prioritize to solve for $\Fd$ by considering it as a hard constraint and add a soft constraint to have a solution as close as possible to $\Md$.

\subsubsection{{QPs for Hyperplanes} (\texorpdfstring{$\text{QP}_{svm}$}{QP\_svm})} \label{subsec:hp_gen_rig}
After solving for $\Fdi$ and $\Fdj$, we use a hybrid soft-hard margin SVM. In particular, we solve for the normal of the hyperplane $\nij$ and an offset $a$.
Since there is no particular desired intersection point between the $\nij$ hyperplane and the rigid payload, $a$ is a decision variable. Accordingly, we propose $\QPsvm$ as
\begin{align}
    &\min_{\nij, a, s_1, s_2} \hspace{0.2cm} \|\mathbf{n}_{ij}\|^2 + \lambda_s(s_1 +s_2)
    \label{eq:svmrig}
    \\
    &\text{\noindent s.t.}\begin{cases}
   \nij^T\opi - a \leq -1, \quad \nij^T\opj - a \geq 1 \\
   \nij^T\pai - a \leq -1, \quad  \nij^T\paj - a \geq 1 \\
   \nij^T(\pai + \Fdi - a) \leq -1 + s_1 \\
   \nij^T(\paj + \Fdj - a) \geq 1 - s_2 \\
    s_1 \geq 0, \quad s_2 \geq 0
    \end{cases}. \nonumber
\end{align}
\begin{figure}[t]
    \centering
    \includegraphics[width=0.486\linewidth]{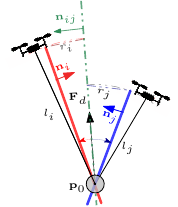}
    \includegraphics[width=0.5\linewidth]{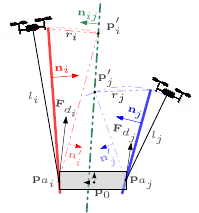}
    \caption{2D projection of the hyperplane manipulation, where $\QPsvm$ computed $\nij$. Left: Point mass case, where $\nij$ is rotated around $\po$ such that the distance to $\nij$ is $r_i$ or $r_j$. Right: Rigid body case, where new intermediate hyperplanes $n_i'$ and $n_j'$ are constructed followed by the same rotation as in the point mass case.}
    \label{fig:2cfs_pm}
\end{figure}

\begin{figure*}[t!]
    \centering
    \includegraphics[width=0.245\textwidth]{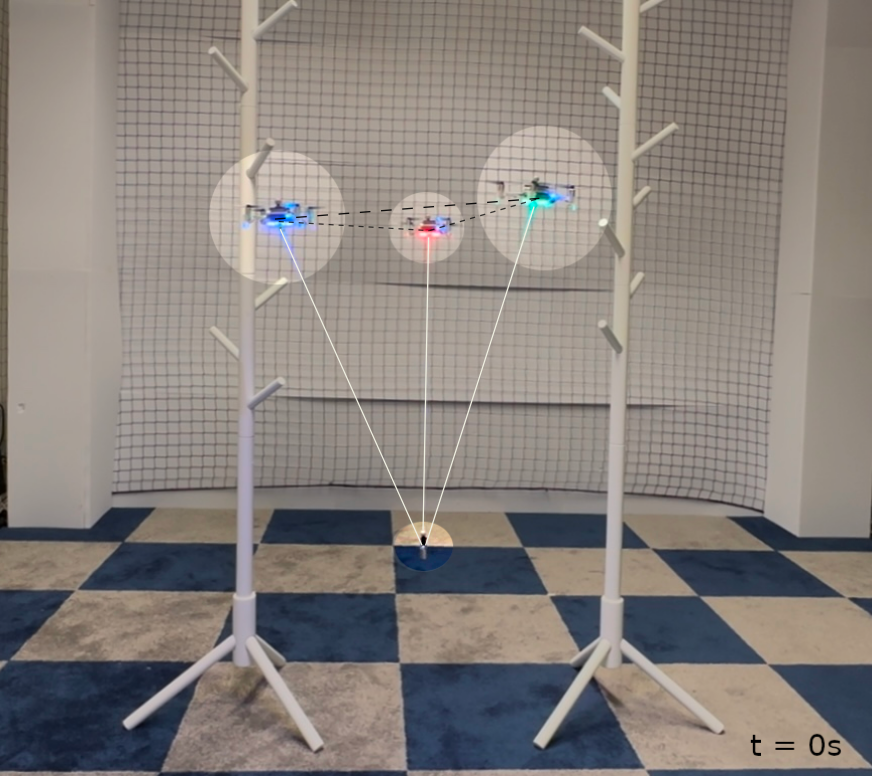}
    \includegraphics[width=0.245\textwidth]{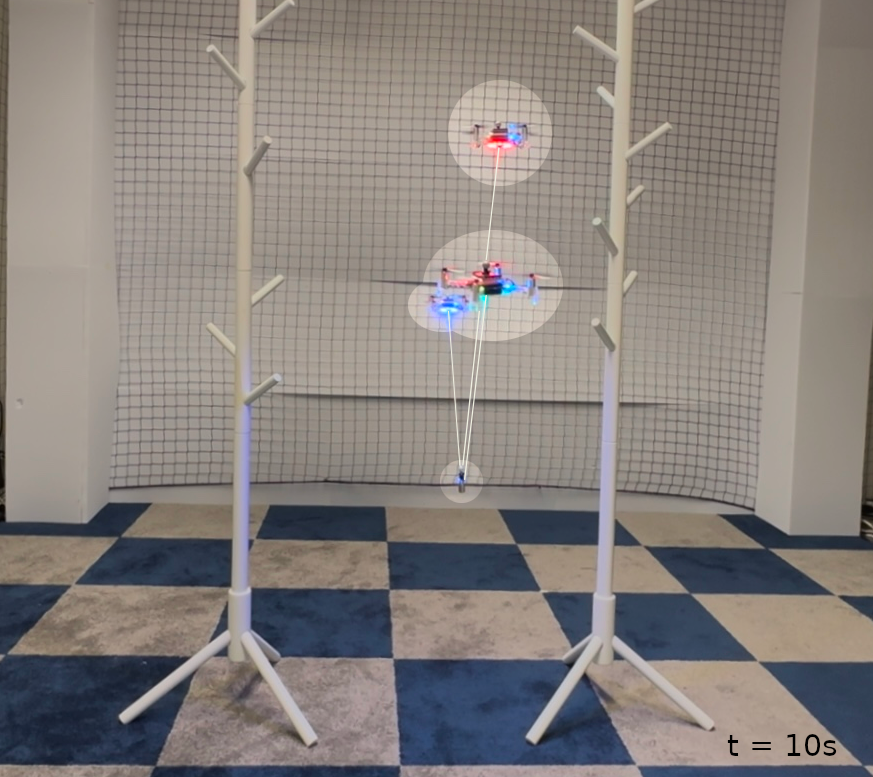}
    \includegraphics[width=0.245\textwidth]{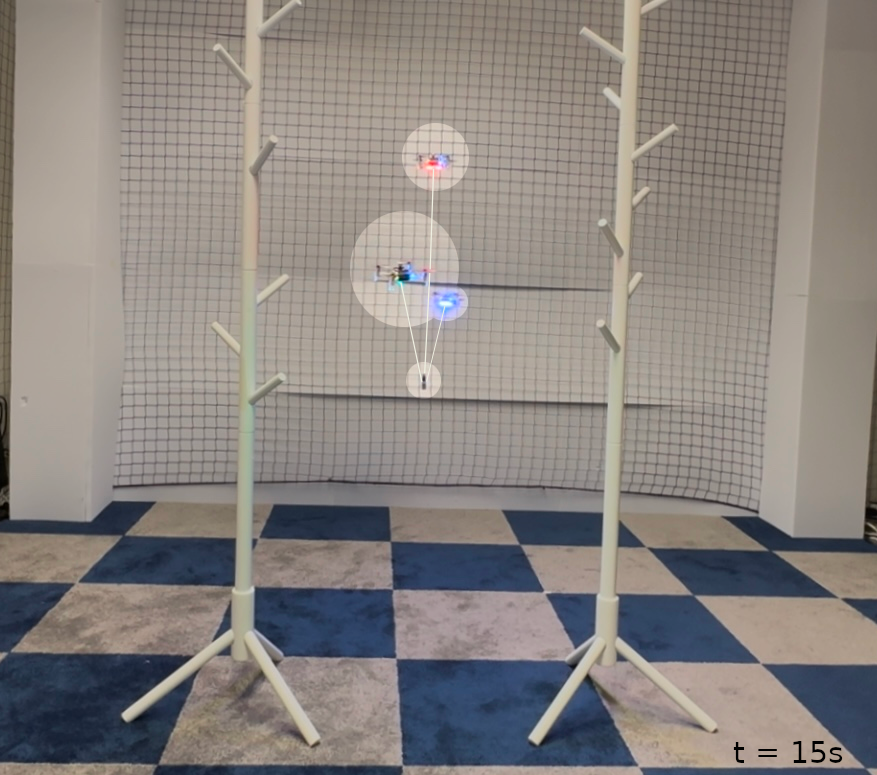}
    \includegraphics[width=0.245\textwidth]{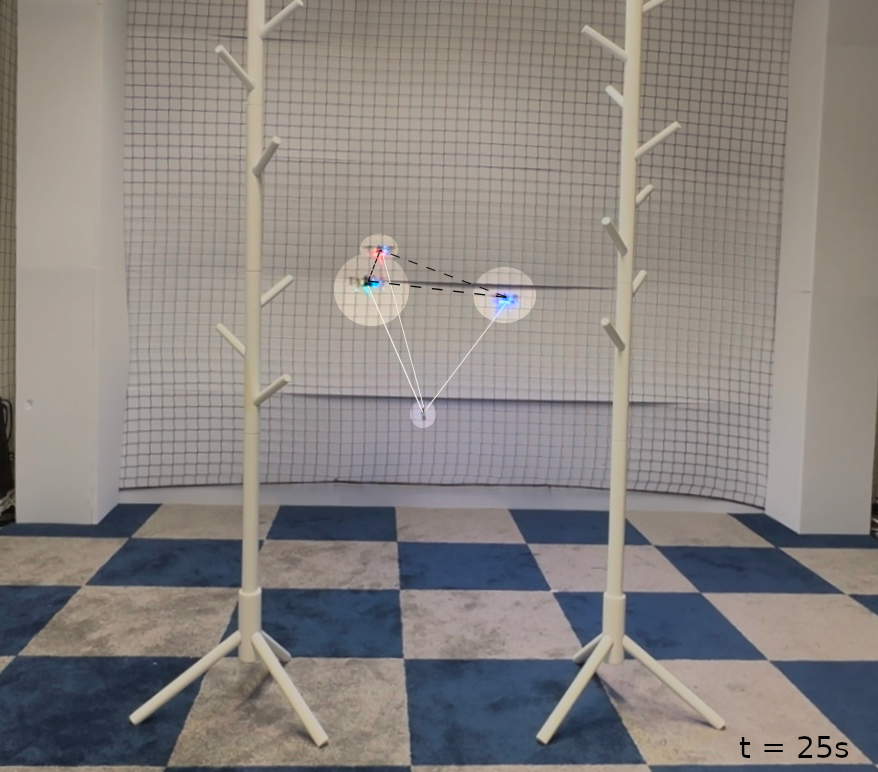}
    \caption{Four frames (manually highlighted for better visibility) of three multirotors carrying a point mass payload in a teleoperation use-case while avoiding inter-robot collisions. There are two obstacles forming a narrow passage. The operator controls the payload position as well as the desired formation. Here, a line configuration as shown in $t = 10 s$ and $t = 15 s$ is used. After passing the obstacles, the operator disables the manual formation control and our approach computes to a more energy-efficient triangle formation ($t = 25 s$).}
    \label{fig:desconfigframes}
\end{figure*}
Here, the first four constraints separate the cables like in a hard-margin SVM. The next four constraints assure that the hyperplane $\nij$ minimizes the projection of both $\Fdi$ and $\Fdj$ after being shifted on the attachment points $\pai, \paj$, respectively. 
We use two slack variables $s_1$ and $s_2$ that are minimized to relax the hard constraints imposed on the projection of $\Fdi$ and $\Fdj$ on the hyperplane of $\nij$. 
This part is similar to a soft-margin SVM.
Similar to the point mass case, $\lambda_s$ also acts as a trade-off between the safety margin and being close to both forces. 
The hyperplane does not have to pass through $\po$ since $a$ is a decision variable.

\subsubsection{Hyperplane Manipulation}

We use the same method for tilting as in the point mass case (\cref{subsec:hyperplane_man_pm}) with extra steps. Let us define two points $\opi^{\prime}$ and $\opj^{\prime}$ as shown in \cref{fig:2cfs_pm} on the right. 
We first compute the intersection between the sphere that describes the motion of each cable and the hyperplane $\nij$, which is a circle on $\nij$.
We use the points of the circle with the highest $z$-coordinate as $\opi^{\prime}$ and $\opj^{\prime}$.
Afterwards, two new hyperplanes are computed as $\nii^{\prime}$ and $\nj^{\prime}$ by using vector $\nij \times \mathbf{e}_3$ and two points for each plane, which are the intersection points $\opi^{\prime}$ and $\opj^{\prime}$ and the attachment points $\pai, \paj$, respectively.
Finally, we apply the same steps from the point mass case to tilt $\nii^{\prime}$ and $\nj^{\prime}$ based on the radii $r_i$ and $r_j$ of each robot model to generate $\nii$ and $\nj$ using \cref{eq:alphas} and \cref{eq:nij}.

\subsection{Desired Cable Forces} \label{sec:desiredcableforces}
One of the main advantages of the $\QPmu$ formulation and in particular the choice of representing the inequality constraints are the extensions that provide new use-cases.
Here, we provide one such example to provide desired cable forces.
There are two motivations behind supporting this feature.
The first motivation is to steer the optimization problem to have consecutive solutions that avoid sudden jumps.
The second motivation is to allow the user to specify a desired formation configuration, for example to pass through a narrow passage.

Let us define $\boldsymbol \mu_{i_{0}}$ as preferable desired cable forces, as either the solution from the previous optimization or the user-specified desired values.
In the latter case, we can rescale $\boldsymbol \mu_{i_{0}}$ such that $\sum_i \boldsymbol \mu_{i_{0}} = \Fd$.
The cost function in $\QPmu$ \cref{eq:newqprigid} can be modified to minimize the difference between the desired cable forces and the preferred ones with a weighting factor $\lambda$ as
\begin{equation}
    c = \frac{1}{2}\normmu^2 + \lambda \| \boldsymbol \mu_{i_{0}} - \mud\|^2.
\end{equation}
\section{Experiments}
We outline our experimental setup and first describe our steps to bridge the sim-to-real gap. For the real platform, we use multirotors of type Bitcraze Crazyflie 2.1 (CF). These are small (\SI{9}{cm} rotor-to-rotor) and lightweight (\SI{34}{g}) products that are commercially available. The physical parameters are identified in prior work~\cite{forster2015system}.
We use the existing extended Kalman filter for state estimation and implement our control algorithm in \emph{C} to run directly on-board the STM32-based flight controller (\SI{168}{MHz}, \SI{192}{kB} RAM).

We use a four-staged \textbf{iterative development method} to reduce challenges regarding the sim-to-real gap.
In the first stage, we implement our experiments in simulation using only \emph{Python}.
For our controller, we write the QP program using an open source Python-embedded modeling language for convex optimization problems, CVXPY~\cite{diamond2016cvxpy}. 
In the second stage, we port our controller from \emph{Python} to \emph{C} and add it to the multirotor's firmware.
We verify the semantic equivalence in our simulator by generating \emph{Python} bindings (Software-in-The-Loop, SITL).
In the third stage, we port our own controller, which requires to solve multiple QPs \cref{eq:newqprigid} on-board the STM32 microcontroller.
The primary challenge is to migrate our QP formulations from CVXPY to \emph{C}, for which we rely on OSQP~\cite{osqp}, which can generate code that is optimized for embedded systems.
As before, we validate this implementation in simulation in a SITL-fashion.
In the fourth stage, we optimize the code to be able to run in realtime on the physical hardware. To this end, we limit ourselves to single-precision floating point operations, solve QPs asynchronously from the control loop, and warm start the optimization with the previous results. 

We \textbf{validate} the functionality of our proposed approach through especially designed physical flight experiments, which highlight that a fatal crash would occur in the absence of our approach that accounts for inter-robot collisions and small in size payloads, while tracking the payload trajectory.

We verify our approach in physical flight tests with up to three Crazyflies.
Each multirotor has its own on-board controller, solving the QPs and using the computed desired cable forces for itself to generate its own control input for the motors.
Such a distributed execution requires that all robots compute identical values.
Since our QPs have a global minimum that the solver reaches, we just need to ensure that all multirotors use the same input data.
On the host side, we use Crazyswarm2,
which is based on Crazyswarm~\cite{preiss2017crazyswarm} but uses ROS~2~\cite{macenski2022robot} to control and send commands for multiple Crazyflies.
Crazyswarm2 heavily relies on broadcast communication for state estimation and control commands, which ensures that all robots receive the data at the same time.
In particular, we equip each multirotor with a single reflective marker for position tracking at $\SI{100}{Hz}$ using an OptiTrack motion capture system.
For the payload, we use a single marker for the point mass or four markers to track the rigid body payload.
Multirotors and payload states are sent with broadcasts to all robots. 
For commands such as takeoff, land, or trajectory execution we also rely on broadcasts to ensure consistent desired states between multirotors.

The states of each multirotor are estimated on-board using an extended Kalman filter and the first derivatives of the payload state using numeric differentiation in combination with a complimentary filter. 
For the payload, we use a calibration weight as point mass, cardboard and 3D-printed objects as rigid bodies, and dental floss as cables.
We use magnets to connect the cables and the payload/multirotors to be easily repaired. 
Our approach does not require measuring or estimating the cable forces, since the controller tracks $\qid$.

We highlight through our simulation results the superiority of our method in scalability and computational efficiency compared to state-of-the-art controllers.

\subsection{Experimental Results}
We conduct several real flight experiments, see \cref{tab:flights}.
For each case, we report the runtime of our optimization (as executed on-board the microcontroller) and the payload pose error.
To the best of our knowledge, existing optimization-based controllers (e.g., \cite{li2023nonlinear, sun2023nonlinear}) that can operate in similar use-cases have not been implemented on a highly constrained computing platform, such as the STM32-based flight controller.
They require more sophisticated solvers, which makes it almost impossible to do so.
Moreover, executing controllers off-board for comparison introduces significant latencies, invalidating the results.
Consequently, a direct performance comparison with such controllers is currently not possible.

We consider two different types of reference motions: a polynomial figure-$8$ trajectory (reaching \SI{0.5}{m/s} for the point mass and \SI{0.4}{m/s} for the rigid body cases), and teleoperation with position and velocity commands for the payload.
For the teleoperation experiment, we put obstacles as shown in \cref{fig:desconfigframes} such that the robots switch to a line configuration (see \cref{sec:desiredcableforces}) to avoid collisions. The operator was able to switch to a predefined configuration by pressing a button.
For the other experiments, we use two types of rigid payloads and different number of Crazyflies and cable lengths.

Overall, we are able to compute desired cable forces at $\SI{30}{Hz}$ on-board the microcontroller, which is sufficient for robust pose tracking of the payload. Surprisingly, the overhead of setting up the QPs takes also a significant amount of time (e.g., \SI{13}{ms} in the 3 UAVs triangle case).
The obtained pose errors are consistent with prior work using bigger multirotors~\cite{liu2022safety}, for positions in the range of the rotor-to-rotor size of the UAV.
The yaw error is higher in the 2 UAVs rod case, because of the challenging nature of our yaw setpoints, that require rotating the formation while flying the figure-8 motion.
\begin{table}[t!]
    \caption{Mean and standard deviation values (small gray) of the control loop runtime in milliseconds.}
    \centering
    \begin{tabular}{l||r|r|r|r}
    Robots & 3 \hspace{3mm} & 6 \hspace{3mm} & 8 \hspace{3mm} & 10 \hspace{2.8mm} \\
    \hline\hline
    Distributed $t_{\textbf{total}}$ & 0.1  {\color{gray}\tiny 0.0} & 1.0 {\color{gray}\tiny 0.2}  & 2.0 {\color{gray}\tiny 0.3}
    &  3.6 {\color{gray}\tiny 0.9}  \\
    \hline
     Centralized $t_{\textbf{c}}$  & 0.1 {\color{gray}\tiny 0.0}   & 3.9 {\color{gray}\tiny 0.6} & 10.9 {\color{gray}\tiny 0.6} & 20.8 {\color{gray}\tiny 3.4}\\
    \hline
    NMPC \cite{sun2023nonlinear}: $t_{\textbf{c}}$& 7.0 \hspace{2.4mm} & 30.0\hspace{3.4mm} & 55.0\hspace{3mm} & 100.0 \hspace{2.2mm}
    \end{tabular}
    \label{tab:scale}
\end{table}
\subsection{Scalability}

To assess the team size scalability, we record the computation time on a laptop (i7-1165G7, \SI{2.8}{GHz}) of the three QPs in \cref{fig:qp_arch} for up to 10 robots. Compared to \cite{sun2023nonlinear},  we report the runtimes that were obtained on a comparable laptop. Our first metric is the total time of the full control loop running in distributed fashion as $t_{\textbf{total}}  = t_{\textbf{svm}} + t_{\Fd} + t_{\mubold} + t_{\textbf{ctrl}}$, where $t_{\textbf{ctrl}}$ is the time of all the other computations.
As shown in \cref{tab:scale}, the distributed time scales roughly quadratically with the number of robots and the overall runtime is at least $1.4$ orders of magnitude faster than the results in \cite{sun2023nonlinear}.

Additionally, since \cite{sun2023nonlinear} is a centralized optimization, we consider the centralized version of our method as the second metric. We compute the time over $n$ robots over the QPs as $t_{\textbf{c}}  = nt_{\textbf{svm}} + nt_{\Fd} + t_{\mubold}$. 
We are still significantly faster than \cite{sun2023nonlinear} with a runtime reduction by a factor of five.
\begin{table*}[t!]
    \caption{Flight Test Results. Shown are mean values over time for a single flight each with standard deviation (small gray).}
    \begin{tabular}{l||c|c|c|c|c}
    & 2 UAVs, point mass & 3 UAVs, point mass & 2 UAVs, rod & 3 UAVs, triangle & 3 UAVs, point mass\\
    \hline\hline
    Payload & point mass & point mass & rod & triangle & point mass\\
    Trajectory & figure 8 (13 s) & figure 8 (13 s) & figure 8 (15 s) & figure 8 (15 s) & teleoperation\\
    Mass [g] & 10 & 10 & 8 & 10 & 10\\
    Dimension [cm] & - & - & 15 & 8 & -\\
    Cables [cm] & 25, 50 & 25, 50, 75 & 50, 50 & 50, 50, 50 & 50, 50, 50\\
    \hline
    QP runtime total [ms] & 5.7 {\color{gray}\tiny 1.5} & 23.1 {\color{gray}\tiny 5.8} & 13.5 {\color{gray}\tiny 2.1} & 33.1 {\color{gray}\tiny 4.3} & 21.8 {\color{gray}\tiny 6.2}\\
    \quad QP runtime Fd [ms] & 0.0 {\color{gray}\tiny 0.0} & 0.0 {\color{gray}\tiny 0.0} & 2.4 {\color{gray}\tiny 0.7} & 6.9 {\color{gray}\tiny 1.3} & 0.0 {\color{gray}\tiny 0.0}\\
    \quad QP runtime SVM [ms] & 2.0 {\color{gray}\tiny 1.2} & 14.7 {\color{gray}\tiny 4.8} & 2.7 {\color{gray}\tiny 1.3} & 8.4 {\color{gray}\tiny 2.9} & 13.0 {\color{gray}\tiny 4.8}\\
    \quad QP runtime $\boldsymbol{\mu}$ [ms] & 1.8 {\color{gray}\tiny 0.8} & 3.9 {\color{gray}\tiny 3.0} & 3.0 {\color{gray}\tiny 1.2} & 4.7 {\color{gray}\tiny 2.7} & 4.4 {\color{gray}\tiny 3.7}\\
    \hline
    Payload $x$, $y$, $z$ error [cm] & 2.5 {\color{gray}\tiny 1.3}, 3.0 {\color{gray}\tiny 2.2}, 2.8 {\color{gray}\tiny 1.4} & 3.6 {\color{gray}\tiny 3.0}, 6.3 {\color{gray}\tiny 3.4}, 4.0 {\color{gray}\tiny 2.3} & 4.3 {\color{gray}\tiny 3.4}, 3.1 {\color{gray}\tiny 2.4}, 4.2 {\color{gray}\tiny 1.7} & 2.4 {\color{gray}\tiny 1.6}, 2.9 {\color{gray}\tiny 1.9}, 3.6 {\color{gray}\tiny 1.6} & 3.0 {\color{gray}\tiny 3.0}, 6.8 {\color{gray}\tiny 4.5}, 5.9 {\color{gray}\tiny 4.1}\\
    Payload $r$, $p$, $y$ error [deg] &  -  &  -  & 6.3 {\color{gray}\tiny 4.5}, 6.9 {\color{gray}\tiny 3.9}, 20.0 {\color{gray}\tiny 7.3} & 10.3 {\color{gray}\tiny 4.3}, 5.9 {\color{gray}\tiny 3.8}, 3.4 {\color{gray}\tiny 1.3} &  - \\
    \end{tabular}
\label{tab:flights}
\end{table*}
\subsection{Challenges}

Implementing the control framework presented in \cref{fig:fullarch} in simulation posed a significant challenge. 
The key difficulty was in identifying the control gains that lie in one of the local optima that is robust enough to balance between the control performance and addressing physical uncertainties upon transfer to the actual platform.
Moreover, executing the full control framework on the physical multirotors in real flights had additional challenges. 
In particular, the geometric controller \cite{lee2017geometric} contained several parts that are numerically unstable for physical flights.
The second loop of the controller computes $\mathbf{u}_i$, which requires either measuring or estimating the acceleration of the payload. The numeric estimate of this value is very noisy and causes the multirotors to crash.
Instead, we rely on the reference acceleration of the payload.
Similarly, the change of the cable unit direction $\qiddot$ is too noisy in practice and we use $\mathbf{0}$ instead.

We also found that tuning the gains of the cascaded design is very challenging, even with access to all relevant tracking errors.
The gains are very sensitive and depend on the number of UAVs and the type of payload.
Moreover, recovering from the disturbances that occur during takeoff poses another challenge especially for tuning.
We note that switching from the single multirotor controller to our controller midflight is a possible alternative, but it is not practical for non-uniform cable lengths.
Some of these challenges might be easier to overcome on platforms with a higher thrust-to-weight ratio than ours with a low ratio of about 1.4, although our data indicates that motors did only occasionally saturate during takeoff in our experiments.
In contrast, the two new gains our method introduces, $\lambda$ and $\lambda_s$, are easy to tune. Both hyperparameters are tuned in our iterative development workflow from sim-to-real in different settings.

\section{Conclusion and Future Work}
We present an efficient optimization-based cable force allocation of a geometric controller for cable-suspended payload transportation that is aware of neighboring robots to avoid collisions.
Unlike previous work that relies on nonlinear optimization, we use a cascaded sequence of small and efficiently solvable quadratic programs.
The stability analysis of the geometric controller still holds since we only operate in the nullspace of the force allocation.
We show that our method scales well with the number of robots with runtime reduction of an order of magnitude compared to the state-of-the-art controllers.
We demonstrate through different physical experiments that our approach can be executed on compute-constrained multirotors in realtime.

An exciting future avenue lies in planning the desired motions and cable forces for time-optimal and collision-free navigation. It would also be an interesting future research direction to analyze the solution quality of our approach with respect to the non-convex formulations. 
Furthermore, a future research direction for the system's model and control is to allow unknown physical parameters of the payload and non-taut cables, rather than the rigid rod assumption.

\printbibliography
\end{document}